%% file: egpaper_for_review.tex
\documentclass[10pt,twocolumn,letterpaper]{article}

\usepackage{iccv}
\usepackage{times}
\usepackage{epsfig}
\usepackage{graphicx}
\usepackage{amsmath}
\usepackage{amssymb}

\usepackage{booktabs}
\usepackage{xcolor}
\definecolor{OliveGreen}{cmyk}{0.64,0,0.95,0.40} 
\definecolor{rred}{cmyk}{0,0.88,0.77,0.26}
\input{shortcuts}
\usepackage[pagebackref=true,breaklinks=true,letterpaper=true,colorlinks,bookmarks=false]{hyperref}

\iccvfinalcopy %

\def\iccvPaperID{36} %

\ificcvfinal\fi

\begin{document}

\title{On Offline Evaluation of 3D Object Detection for Autonomous Driving}

\author{
Tim Schreier \quad Katrin Renz \quad Andreas Geiger \quad Kashyap Chitta \\
University of Tübingen \quad Tübingen AI Center\\
{\tt\small tschreier2@gmail.com \quad \{katrin.renz, a.geiger, kashyap.chitta\}@uni-tuebingen.de}
}

\maketitle
\ificcvfinal\thispagestyle{empty}\fi

\begin{abstract}
Prior work in 3D object detection evaluates models using offline metrics like average precision since closed-loop online evaluation on the downstream driving task is costly. However, it is unclear how indicative offline results are of driving performance.
In this work, we perform the first empirical evaluation measuring how predictive different detection metrics are of driving performance when detectors are integrated into a full self-driving stack. We conduct extensive experiments on urban driving in the CARLA simulator using 16 object detection models. We find that the nuScenes Detection Score has a higher correlation to driving performance than the widely used average precision metric. In addition, our results call for caution on the exclusive reliance on the emerging class of `planner-centric' metrics. 
\end{abstract}
\vspace{-0.8cm}
\vspace{-0.09cm}
\section{Introduction}

Ever since the first object detection benchmark challenges like the PASCAL VOC \cite{everingham20062005} became popular, mean average precision (mAP) has been used as the standard metric for evaluating the performance of detection models. 
Recent works \cite{nds,sde,injecting} have criticized mAP for its task-agnostic design as the metric assigns equal importance to all objects, which does not reflect real-world priorities for self-driving.
Therefore, different task-specific modifications of mAP \cite{sde,nuscenes,kitti,waymodata} and planner-centric approaches to detection evaluation \cite{nds,injecting,tip} have been proposed.
These offline metrics are useful because they are quick, cheap, and safe to evaluate compared to online tests. However, relying solely on offline evaluation is only useful if it strongly correlates with the actual driving performance.
With the influx of new detection metrics, it has become unclear which metric researchers should rely on and how the metrics compare.

\begin{figure}
\centering
\includegraphics[width=0.48\textwidth]{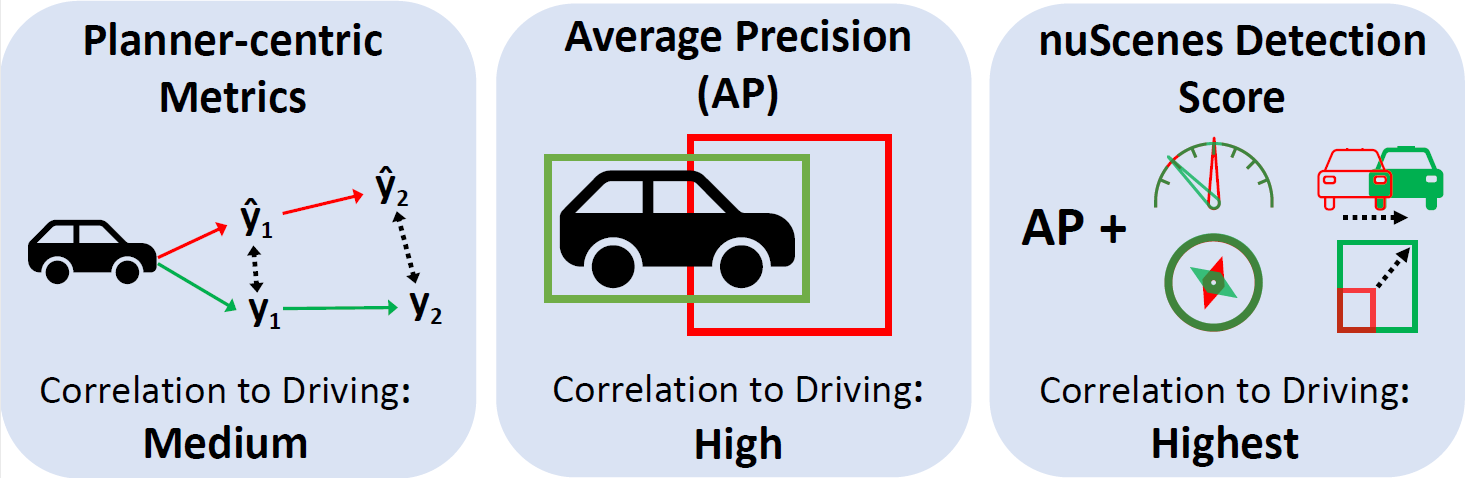}
\caption{\textbf{Summary of findings.}
nuScenes Detection Score is most predictive of driving performance in an extensive empirical study on the CARLA simulator.
}
\vspace{-0.25cm}
\label{fig:MAP}
\end{figure}

In this work\footnote {To read the full-length report, please visit:  \href{https://t.ly/CsIrt}{https://t.ly/CsIrt}.}, we provide the first empirical evidence of how predictive detection metrics are of downstream driving performance. We train $16$ modern 3D detectors, integrate them into a self-driving pipeline, and evaluate their performance in the CARLA simulator \cite{carla}. This allows us to study how strongly these metrics correlate with driving outcomes.
We find that even though mAP is highly correlated with driving performance, the nuScenes Detection Score \cite{nuscenes}, a task-specific variation, is even more predictive. Furthermore, the planner-centric metrics we examine, which measure the impact of inaccurate detections on planner outcomes, are significantly less indicative of driving performance. Our key findings are summarized in \figref{fig:MAP}.
\vspace{-0.09cm}

\section{Related Work}\label{sec:relatedwork}

\boldparagraph{Task-specific Detection Metrics}\label{sec:perf-metrics}
To track algorithmic advancements, researchers compare object detection models on dataset-based competitions~\cite{nuscenes,datas1,waymodata,kesten2019lyft,kitti} using the mAP metric \cite{3ds1}, which does not take the egocentric nature and task-specific characteristics of driving into account. Therefore, prior work has proposed task-specific object detection metrics for self-driving~\cite{nds,waymodata,tip,rrr,injecting}. One approach is taking mAP and adapting it to self-driving.
mAPH, the principal metric of the Waymo challenge \cite{waymodata} and the Average Orientation Similarity (AOS) \cite{kitti} are both designed to account for the importance of correct heading estimation for behaviour planning and weigh detections accordingly.
Similarly, Deng et al. \cite{sde} suggest evaluating detections from an egocentric perspective and introduce the Support Distance Error (SDE). 

Another approach focuses on evaluating the effects of perception errors on the planning module. The planning-KL-divergence (PKL) \cite{pkl} measures the KL-divergences between distributions of waypoint locations conditioned either on noisy perception or ground truth object annotations.
Li et al. \cite{tip} emphasize the importance of understanding the internal reasoning of a planner, as some detection failures do not cause immediate behaviour change. However, their approach focuses on model-based planners and does not apply to neural architectures.
Ivanovic and Pavone \cite{injecting} describe an approach where local gradients of a hand-crafted planning function are used to assign object weights. 

None of these prior works have provided quantitative results demonstrating that the metric they propose is more closely related to measures of driving performance than the standard mAP metric. We seek to address this gap in the literature by comparing offline evaluations with online tests.

\boldparagraph{Online and Offline Evaluation}\label{online-offline}
Deep neural networks for self-driving applications are commonly first tested offline with the use of a pre-recorded dataset~\cite{riccio2020testing}.
Offline results are used as proxy performance indicators for online evaluations, which are more expensive to conduct. 
It is often unclear how offline measurements relate to system-level functionality in embedded systems. Haq et al. \cite{ex1} evaluate the correlations between online and offline performance of a camera-based end-to-end lane-keeping model.
The authors conclude that offline evaluations cannot be used for safety testing in the context of steering prediction. However, a recent replication of this study with improved methodology obtains a tighter relationship between online and offline results \cite{stocco2023model}.
In similar experiments, prior work has found little to no correlation when comparing online driving performance to offline prediction accuracy for agents in the CARLA \cite{ex2} and nuPlan \cite{Dauner2023ARXIV} simulators. %

Contrary to these efforts, we do not focus on steering prediction but aim to evaluate the effects of detection errors on driving outcomes. 
We aim to provide the first quantitative evidence comparing detection metrics for self-driving to help the community choose relevant metrics.

\vspace{-0.09cm}
\section{Modular Pipeline for Autonomous Driving}
\label{Methods} 

In this section, we present our driving agent. We introduce the problem setting and describe the modular structure of our agent. Next, we discuss the modules for object detection, tracking, and motion planning.

\boldparagraph{Task and Setup}\label{sec:problem-setting}
We address the task of urban driving with the objective of navigating safely along a predetermined route while adhering to traffic regulations. The agent predicts steering and throttle from sensor inputs. While all traffic participants need to be recognized via a LiDAR-based object detector, our agent has privileged access to simulator information concerning the ego lane and traffic light states.

\boldparagraph{Pipeline}
At every timestep,
the point cloud of a $360^{\circ}$ LiDAR sensor is processed by a 3D object detector which produces a set of oriented bounding box predictions of traffic participants in the scene.
Detections are tracked over time
to yield consistent predictions and speed estimates for detected objects. Given its high-level navigation goal, the planner then uses these predictions to decide on an appropriate set of target waypoints that encode the planned trajectory. A PID controller then processes the waypoints to compute appropriate lateral and longitudinal controls.
To study the effects of the performance of specific object detectors on the downstream task of driving, all pipeline elements except the detector are constant across experiments. 

\boldparagraph{Detection and Tracking}
For our experiments, we consider eight different LiDAR-based 3D object detection architectures. To cover a breadth of architectures in our analysis, we include voxel-based detectors \cite{vrcnn,second}, a pillar-based detector \cite{pillarnet}, and approaches that also incorporate point-based information from the point cloud \cite{pointpillars,pv,pv++,pv,parta2}. This array of detection architectures further includes a mix of anchor-based and anchor-free detection heads as well as single-stage and two-stage approaches. In our setup, detections are associated across frames using Hungarian matching \cite{kuhn1955hungarian}. The speed of tracked objects is approximated via a simple heuristic: per object track, the bounding box centers of the last two timesteps are projected to the ground plane. The L2 norm between these points is divided by the timestep length to approximate speed.

\boldparagraph{Planning}
For our experiments, we choose PlanT \cite{plant} for motion planning. PlanT is a transformer-based planner with state-of-the-art performance in the CARLA simulator.
\vspace{-0.09cm}
\section{Metrics}
\label{Metrics} 

This section introduces the metrics we use in our experiments: the online metrics that quantify driving performance, the mAP metric, its task-specific variations, and finally two planner-centric metrics.

\boldparagraph{Online Metrics}
\label{onlinemetrics}
Online evaluations provide the most reliable estimates of system-level performance. In this work, we use the CARLA Driving Score, the official metric for the CARLA leaderboard \cite{cleader}, and the number of collisions.

\noindent \textit{Driving Score.}
The CARLA Driving Score (DS) is a composite metric combining route completion with the infraction score.
The route completion (RC) describes the percentage of the route the agent completed and the infraction score (IS) measures collisions or violations of traffic rules.

\noindent \textit{Collision Count.}
As a second metric, we count the total number of collisions per evaluation (\#Col.) in which the ego vehicle is involved. This metric exclusively focuses on safety since it does not depend on the route completion.

\boldparagraph{Average Precision based Metrics}
\label{sec:map}
\textit{Average precision} is the standard metric to measure performance on detection tasks. 
It is defined as the area under the precision-recall curve.
Following the KITTI protocol \cite{kitti}, we use an intersection over union (IoU) threshold of $70$\% as the true positive criterion and compute the integral using a 40-point interpolation with equidistant recall values.

\noindent \textit{Average Orientation Similarity.}
\label{aos} As the original mAP metric does not account for a notion of heading, we have included the average orientation similarity metric (AOS) \cite{kitti} in our analysis. AOS modifies AP by weighting true positive detections according to the accuracy of their heading predictions. The heading angles of vehicles in the scene provide essential information for motion forecasting and behavior planning \cite{kitti,nuscenes,waymodata}. %

\noindent \textit{Inverse Distance weighted AP.}
\label{inv_map}
Another approach for modifying AP is weighting detections by their inverse distance to the ego vehicle \cite{sde}. Closer objects are inherently more safety-critical than those far away. Thus, one should expect the weighting of AP by the inverse distance (ID-AP) to produce a metric that better reflects the ego-centric nature of detection for self-driving. 

\boldparagraph{nuScenes Detection Score}
The \textit{nuScenes Detection Score} (NDS) \cite{nuscenes} is a popular task-specific detection metric that uses mAP as a starting point. It relies on center distance in birds-eye view (BEV) for the true positive criterion instead of the standard IoU-based approach. 
Here, we use a fixed center distance of 1 meter to not confound our results by averaging across thresholds (the authors suggest averaging over four thresholds).
By relying on center distances, the authors decouple mAP from object size and orientation, which they account for separately. They argue that center distance covers objects of different sizes more evenly because smaller volume objects like pedestrians can quickly achieve an IoU of zero if predictions have minor translation errors.
NDS also includes five explicit detection quality measures for all true positive detections. These measures are weighted equally and are, in total, given as much weight as average precision in the NDS. The five true positive metrics are (1) Average Translation Error (ATE): Euclidean center distance in BEV in meters; (2) Average Scale Error (ASE): Calculated as 1 - IOU after aligning centers and orientation; (3) Average Orientation Error (AOE): The smaller yaw angle between GT and prediction in radians; (4) Average Velocity Error (AVE): Absolute velocity error in m/s; and (5) Average Attribute Error (AAE): (1-Acc) for the prediction accuracy of additional attributes. We do not include the AAE in our experiments, as the additional attributes are specific to the nuScenes dataset.

Similar to mAP, NDS does not account for an object's distance to the ego vehicle. We thus also test an inverse distance-weighted version of the NDS (ID-NDS).

\boldparagraph{Planner-Centric Detection Metrics} \label{planner-centric}
Several authors have recently pursued a novel approach to detection metrics for self-driving: focusing on the planner instead of evaluating object detections directly \cite{pkl,tip,injecting}. To evaluate the planner-centric metrics, we first compute the planner's output for a scene given the ground truth object information. We then also calculate the planner's output given the noisy output of the perception stack (detection + tracking) and analyse how the two predicted trajectories (i.e., the sets of waypoints) differ. This approach naturally down-weights distant and less relevant objects in the metrics as they bear little significance to planning.

We use the average and final displacement errors \cite{nuplan} to quantify differences in trajectories.
The \textit{average displacement error} (ADE) at timestep $t$ is the average of the point-wise L2 distances between the predicted waypoints based on ground truth detections and the predicted waypoints based on noisy detections. We define the ADE for a route as the mean of all frame-based ADEs in that route. Similarly, we define the \textit{final displacement error} (FDE) as the L2 distance between the final waypoints of two trajectories, averaged over all frames of a route.

\vspace{-0.09cm}
\section{Experiments}\label{Experiments} 

This section discusses our setup for online evaluation and explains how we train and configure the object detection models. We then present the results of our analysis.

\boldparagraph{Online Evaluation}
We base our experiments on the CARLA simulator (Version: 0.9.10) \cite{carla}. 
Within CARLA, we gauge online performance for the detection models by integrating them into our modular pipeline to test driving performance. 
We use the Longest6 benchmark \cite{tf} as an online evaluation protocol. Longest6 contains $36$ $\sim$1.5km long routes with high traffic density across six towns. %

While evaluating the agents on Longest6, we log detailed ground truth bounding box information about the objects in the scene, the perception stack's predictions, and the sensor information.  
Based on the resulting logs, we compute the offline detection metrics for every route. This enables us to compare the observed driving outcomes for a given route with the associated offline detection performance measures.  

\boldparagraph{Implementation}
We train eight detection architectures using the open-source LiDAR detection framework OpenPCDet \cite{openpcdet} and the default hyperparameter configurations it provides.
All architectures are trained for $72$ hours on eight NVIDIA GeForce RTX $2080$TIs.
We include two checkpoints per architecture to increase the number of data points in our analysis. To ensure variance in checkpoints' behavior, the first is extracted after only $36$ hours of training and the second after $72$ hours. In total, this results in $16$ models.

We generate the training data via the CARLA simulator by observing the driving behaviour of an expert algorithm \cite{tf} with privileged access to ground truth information. %
At every frame, we record $360^{\circ}$ LiDAR data with a rotation frequency of $20$ Hz.
We set the minimum confidence threshold for detections to $0.3$, apply non-maximum suppression with an IoU threshold of $0.2$ and only present object tracks to the planner that are tracked for four consecutive frames.

\boldparagraph{Metric Evaluation Protocol} After we evaluate the driving performance for all 16 models on the Longest6 benchmark, we average scores across the benchmark's 36 routes to attain one data point per detector for every metric. For the 16 detector-wise data points, we then calculate Pearson's correlation coefficients between offline and online metrics. For ease of comparison and interpretation, we provide all Pearson's $r$ correlation coefficients as absolute values. 

\boldparagraph{Correlation Results}
\begin{table}
\footnotesize
\centering
\setlength{\tabcolsep}{4pt}
\begin{tabular}{lcc @{}}

\textbf{Metric} & \textbf{Correlation: DS} &  \textbf{Correlation: \#Col. }\\
\midrule

nuScenes Detection Score \cite{nuscenes}        &             \textcolor{OliveGreen}{$0.852$} & \textcolor{OliveGreen}{$0.907$} \\

Average Precision \cite{hoiem2009pascal}       &             $0.805$ &                     $0.903$ \\

Avg. Displacement Error \cite{nuplan}        &             $0.784$ &                     $0.770$ \\
Avg. Orientation Similarity \cite{kitti}       &             $0.742$ &                     $0.894$ \\
Final Displacement Error \cite{nuplan}      &             \textcolor{rred}{$0.703$} &             \textcolor{rred}{$0.653$} \\
\bottomrule
\vspace{0.05cm}
\end{tabular}
\caption{\label{table:cor}\textbf{Pearson correlation between online and offline metrics}. Absolute values are shown for clarity.}
\vspace{-0.25cm}
\end{table}
Table \ref{table:cor} presents the correlation coefficients for the base metrics. Note that we only discuss Pearson correlation here for the sake of conciseness but found very similar relations when computing Spearman coefficients. 
The first clear result from the analysis is that all offline metrics correlate highly to driving outcomes. NDS achieves the highest correlation values, with the standard mAP value in second place. The planner-centric metrics achieve less impressive results than their AP-based alternatives. 
The strength of the correlation indicates that offline metrics can provide reasonable heuristics for a detector's performance in online tests. Offline metrics thus seem to be a reliable proxy for rougher comparisons among models and quick hypotheses testing. This insight is significant as the computational cost for online tests is high. The offline metrics only take around four minutes to evaluate, while testing a single model on Longest6 using an NVIDIA GeForce RTX $2080$Ti takes three days.

\boldparagraph{nuScenes Detection Score Ablation}
Table \ref{table:cor} shows that NDS correlates more to the CARLA Driving Score than the standard mAP metric ($0.85$ vs. $0.80$). Both metrics have similar correlations to the number of collisions ($0.90$). 

In the original NDS metric, all detection quality measures are given equal importance, and their sum is given as much weight as mAP. 
We find that even though all detection quality metrics contribute a little bit toward the strong correlation, switching all of them off also produces a metric that correlates more with the Driving Score than mAP  ($0.82$ vs. $0.80$). This suggests that the center distance approach is indeed preferable to an IoU based TP-criterion.

\boldparagraph{Inverse Distance Weighting}
\begin{figure}
\centering
\includegraphics[width=0.48\textwidth]
{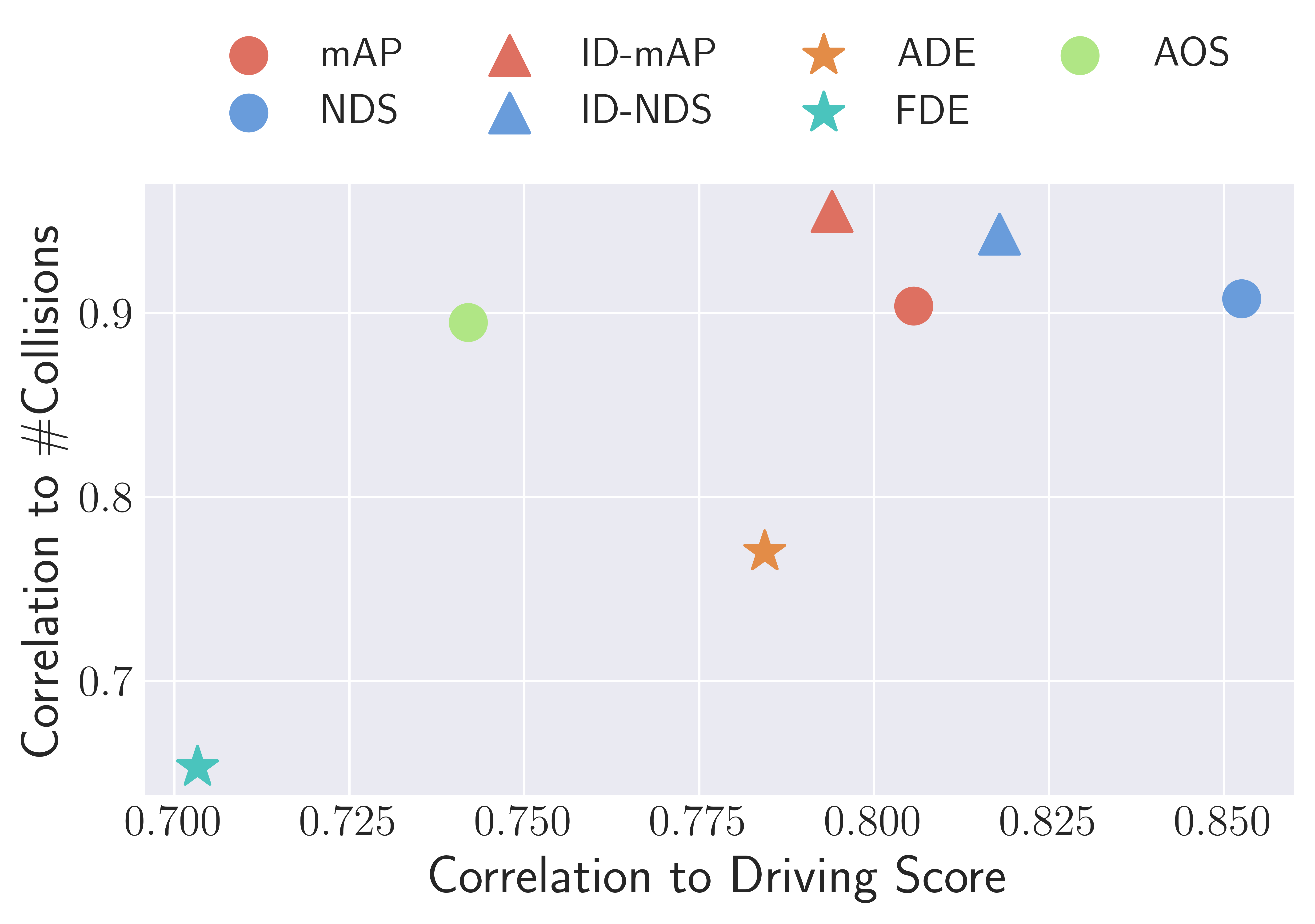}
\caption{\textbf{Correlation analysis summary.} The plot marks correlation coefficients for planner-centric metrics with a star symbol. The inverse distance weighted metrics are marked with triangles.}
\label{fig:PC}
\vspace{-0.25cm}
\end{figure}
We include inverse distance weighted variations for mAP and NDS. We observe that the correlation to Driving Score drops for both metrics (comparing triangles with circles in Figure~\ref{fig:PC}). 
In contrast, the correlation to the number of collisions increases compared to the original metrics. The fact that inverse distance weighting increases predictive power for collisions is expected. Many collisions likely occur when the perception stack misses traffic participants directly in front of the ego. 
However, inverse distance weighting rewards defensive agents, that achieve less route completion.
ID-MAP is especially tightly connected to the collision count, with a Pearson's $r$ coefficient of $0.955$. 

\boldparagraph{Planner-Centric Metrics}
Our results show that ADE marks a better indicator for driving performance than FDE. While the planner-centric ADE does correlate to Driving Score and collision count ($0.78$ \& $0.77$, respectively), these correlations pale in comparison to those of the mAP-based metrics (see Figure \ref{fig:PC}). The significant discrepancies in correlations we observe between the approaches, therefore, contra-indicate a reliance on planner-centric metrics alone when evaluating object detection for self-driving.

\vspace{-0.09cm}

\section{Conclusion}

In this work, we demonstrate that common metrics for 3D object detection are highly correlated with online driving performance.
Our extensive evaluation shows that the nuScenes Detection Score is more predictive of closed-loop outcomes than the standard mean average precision. While we find the standard mAP score to yield strong correlation nonetheless, our results invoke skepticism regarding detection benchmarks that exclusively rely on planner-centric approaches or use a strong focus on heading accuracy. 

There are two important limitations that constrain the degree to which one can generalize from our results. First, we base all our experiments on the same neural planning architecture. In our experiments, PlanT acts as a mediator between the detection performance and the driving outcomes. Different planners might focus on other object cues for motion forecasting and behavior planning. 
Second, the online metric Driving Score is a relatively simple heuristic for evaluating overall driving performance, and more accurate online metrics might yield different correlation outcomes.

\clearpage
\boldparagraph{Acknowledgements} This work was supported by the BMBF (Tübingen AI Center, FKZ: 01IS18039A), the DFG (SFB 1233, TP 17, project number: 276693517), and by EXC (number 2064/1 – project number 390727645). We thank the International Max Planck Research School for Intelligent Systems (IMPRS-IS) for supporting K. Renz and K. Chitta.  The authors also thank Luis Winckelmann for several helpful discussions.

{\small
\bibliographystyle{ieee_fullname}
\bibliography{egbib}
}

\end{document}

%% file: shortcuts.tex
\newcommand{\figref}[1]{Fig.~\ref{#1}}

\makeatother

\newcommand{\boldparagraph}[1]{\vspace{0.1cm}\noindent{\bf #1.}}

\definecolor{darkgreen}{rgb}{0,0.7,0}
\definecolor{darkyellow}{rgb}{0.8,0.8,0}
\definecolor{bittersweet}{rgb}{1.0, 0.44, 0.37}
\definecolor{amber}{rgb}{1.0, 0.49, 0.0}
\definecolor{lgray}{rgb}{0.83,0.83,0.83}

\definecolor{color_unlabled}{rgb}{0.0,0.0,0.0}
\definecolor{color_vehicle}{rgb}{0.0,0.0,0.56}
\definecolor{color_road}{rgb}{0.5,0.25,0.5}
\definecolor{color_redlight}{rgb}{1.0,0.0,0.0}
\definecolor{color_person}{rgb}{0.859,0.078,0.234}
\definecolor{color_roadline}{rgb}{0.613,0.914,0.195}
\definecolor{color_sidewalk}{rgb}{0.953,0.137,0.906}

\definecolor{ellisred}{rgb}{0.87,0.44,0.38} %
\definecolor{ellisgreen}{rgb}{0.69,0.90,0.52} %
\definecolor{elliscyan}{rgb}{0.29,0.77,0.74} %
\definecolor{ellisorange}{rgb}{0.89,0.55,0.28} %
\definecolor{ellisblue}{rgb}{0.41,0.61,0.86} %